\pdfoutput=1
\documentclass[11pt]{article}

\usepackage{acl}

\usepackage{times}
\usepackage{latexsym}

\usepackage[T1]{fontenc}
\usepackage[utf8]{inputenc}

\usepackage{microtype}

\usepackage{amsmath,amsfonts,amsthm,amssymb,bbm}
\usepackage{graphicx}
\usepackage{mathtools}
\usepackage{subfigure}
\usepackage{enumitem}
\usepackage{booktabs, multirow, bigdelim}
\usepackage[group-separator={,}]{siunitx}
\usepackage{soul}           % for strikethrough
\usepackage{numprint}
\npthousandsep{,}

\usepackage{balance}

\usepackage{pifont}% http://ctan.org/pkg/pifont
\newcommand{\cmark}{\ding{51}}%

\usepackage{array}
\newcolumntype{L}[1]{>{\raggedright\let\newline\\\arraybackslash\hspace{-0.35cm}}m{#1}}

\usepackage[ruled,noend,algo2e]{algorithm2e}
\SetKwInput{KwInput}{input}                % Set the Input
\SetKwInput{KwOutput}{output}              % set the Output

\makeatletter
\newcommand{\removelatexerror}{\let\@latex@error\@gobble}
\makeatother
\title{SAIS: Supervising and Augmenting Intermediate Steps for Document-Level Relation Extraction}

\author{
    Yuxin Xiao$^{1}$, Zecheng Zhang$^{2}$, Yuning Mao$^{3}$, Carl Yang$^{4}$, Jiawei Han$^{3}$ \\
    $^{1}$Carnegie Mellon University, $^{2}$Stanford University,\\
    $^{3}$University of Illinois at Urbana-Champaign, $^{4}$Emory University\\
    \texttt{$^{1}$yuxinxia@cs.cmu.edu $^{2}$zecheng@cs.stanford.edu} \\
    \texttt{$^{3}$\{yuningm2,hanj\}@illinois.edu $^{4}$j.carlyang@emory.edu}
}

\begin{document}
\maketitle

\begin{abstract}
Stepping from sentence-level to document-level, the research on relation extraction (RE) confronts increasing text length and more complicated entity interactions.
Consequently, it is more challenging to encode the key information sources---relevant contexts and entity types.
However, existing methods only \textit{implicitly} learn to model these critical information sources while being trained for RE.
As a result, they suffer the problems of ineffective supervision and uninterpretable model predictions.
In contrast, we propose to \textit{explicitly} teach the model to capture relevant contexts and entity types by \underline{S}upervising and \underline{A}ugmenting \underline{I}ntermediate \underline{S}teps (SAIS) for RE.
Based on a broad spectrum of carefully designed tasks, our proposed SAIS method not only extracts relations of better quality due to more effective supervision, but also retrieves the corresponding supporting evidence more accurately so as to enhance interpretability.
By assessing model uncertainty, SAIS further boosts the performance via evidence-based data augmentation and ensemble inference while reducing the computational cost.
Eventually, SAIS delivers state-of-the-art RE results on three benchmarks (DocRED, CDR, and GDA) and
outperforms the runner-up by $5.04\%$ relatively in F1 score in evidence retrieval on DocRED.\footnote{Our code is available at \url{https://github.com/xiaoyuxin1002/SAIS}.}
\end{abstract}
\section{Introduction} \label{sec:1}

Playing a crucial role in the continuing effort of transforming unstructured text into structured knowledge, RE \cite{bach2007review} seeks to identify relations between an entity pair based on a given piece of text.
Earlier studies mostly pay attention to sentence-level RE \cite{zhang2017position, hendrickx2019semeval} (i.e., the targeting entity pair co-occur within a sentence) and achieve promising results \cite{zhang2019long, zhou2020nero}.
Based on an extensive empirical analysis, \citet{peng2020learning} reveals that textual contexts and entity types are the major information sources that lead to the success of prior approaches.

Given that more complicated relations are often expressed by multiple sentences, recent focus of RE has been largely shifted to the document level \cite{yao2019docred, cheng2021hacred}.
Existing document-level RE methods \cite{zeng2020double, zhou2021document} utilize advanced neural architectures such as heterogeneous graph neural networks \cite{yang2020heterogeneous} and pre-trained language models \cite{xu2021pre}.
However, although documents typically include longer contexts and more intricate entity interactions, most prior methods only \textit{implicitly} learn to encode contexts and entity types while being trained for RE.
As a result, they deliver inferior and uninterpretable results.

\begin{figure*}[t!]
    \centering
    \includegraphics[width=\textwidth]{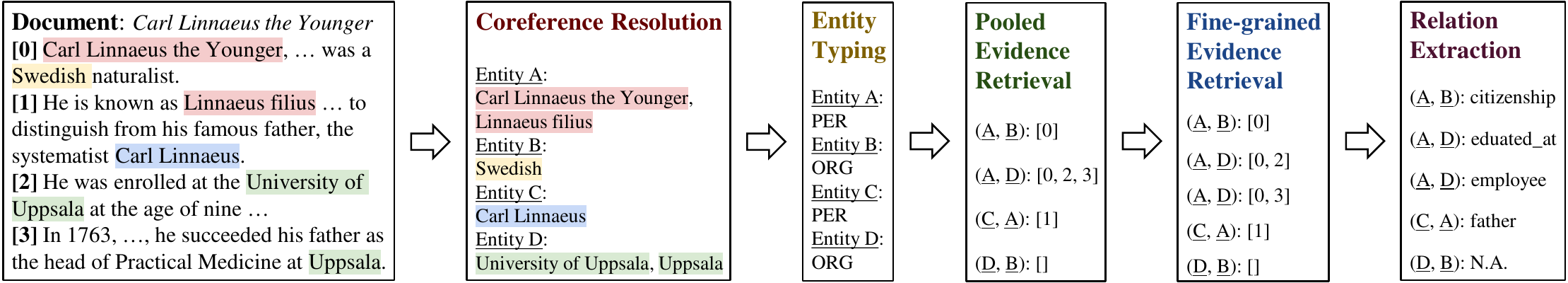}
    \caption{
    Motivating example adapted from DocRED. 
    From the input document with annotated entity mentions to the RE output, there are four intermediate steps involved in the reasoning process.
    These steps are complementary to RE, in the sense that CR, PER, and FER capture textual contexts while ET preserves entity type information.}
    \label{fig:1}
\end{figure*}

% On the other hand, many recent datasets tend to support multiple tasks for the training of more powerful language models.
% For instance, the three popular document-level RE benchmarks used in our experiments allow additional tasks by providing coreference and evidence annotations.
On the other hand, it has been a trend that many recent datasets support the training of more powerful language models by providing multi-task annotations such as coreference and evidence \cite{yao2019docred, li2016biocreative, wu2019renet}.
Therefore, in contrast to existing methods, we advocate for \textit{explicitly} guiding the model to capture textual contexts and entity type information by \underline{S}upervising and \underline{A}ugmenting \underline{I}ntermediate \underline{S}teps (SAIS) for RE.
More specifically, we argue that, from the input document with annotated entity mentions to the ultimate output of RE, there are four intermediate steps involved in the reasoning process.
Consider the motivating example in Figure~\ref{fig:1}:
\begin{description}[style=unboxed,nolistsep]
    \item [\textnormal{(1)} Coreference Resolution (CR):] 
    Although Sentence 0 describes the ``\textit{citizenship}" of ``\textit{Carl Linnaeus the Younger}" and Sentence 1 discusses the ``\textit{father}" of ``\textit{Linnaeus filius}", the two names essentially refer to the same person.
    Hence, given a document, we need to first resolve various contextual roles represented by different mentions of the same entity via CR.
    \item [\textnormal{(2)} Entity Typing (ET):] 
    After gathering contextual information from entity mentions, ET regularizes entity representations with the corresponding type information (e.g., Entity A, ``\textit{Linnaeus filius}", is of type ``\textit{PER}" (person)).
    Within an entity pair, the type information of the head and tail entities can be used to filter out impossible relations, as the relation ``\textit{year\_of\_birth}" can never appear between two entities of type ``\textit{PER}",  for instance.
    \item [\textnormal{(3)} Pooled \textnormal{and} \textnormal{(4)} Fine-grained Evidence Retrieval (PER \textnormal{and} FER):] 
    A unique task for locating the relevant contexts within a document for an entity pair with any valid relation is to retrieve the evidence sentences supporting the relation.
    Nonetheless, some entity pairs may not express valid relations within the given document (e.g., Entities D and B in the example). Meanwhile some entity pairs possess multiple relations (e.g., Entity A is both ``\textit{educated\_at}" and an ``\textit{employee}" of Entity D), each with a different evidence set.
    Therefore, we use PER to distinguish entity pairs with and without valid supporting sentences and FER to output more interpretable evidence unique to each valid relation of an entity pair.
\end{description}
In this way, the four intermediate steps are \textit{complementary} to RE, in the sense that CR, PER, and FER capture textual contexts while ET preserves entity type information. 
Consequently, by explicitly supervising the model's outputs in these intermediate steps via carefully designed tasks, we extract relations of improved quality.

In addition, based on the predicted evidence, we filtrate relevant contexts by augmenting specific intermediate steps with pseudo documents or attention masks.
By assessing model confidence, we apply these two kinds of evidence-based data augmentation together with ensemble inference, only when the model is \textit{uncertain} about its original predictions.
Eventually, we further boost the performance with negligible computational cost.

Altogether, our SAIS method achieves state-of-the-art RE performance on three benchmarks (DocRED \cite{yao2019docred}, CDR \cite{li2016biocreative}, and GDA \cite{wu2019renet}) due to more effective supervision and enhances interpretability by improving the evidence retrieval (ER) F1 score on DocRED by $5.04\%$ relatively compared to the runner-up.
\section{Background} \label{sec:2}

\subsection{Problem Formulation} \label{sec:2.1}

Consider a document $d$ containing sentences $\mathcal{S}_d = \{s_i\}_{i=1}^{|\mathcal{S}_d|}$ and entities $\mathcal{E}_d = \{e_i\}_{i=1}^{|\mathcal{E}_d|}$ where each entity $e$ is assigned an entity type $c \in \mathcal{C}$ and appears at least once in $d$ by its mentions $\mathcal{M}_e = \{m_i\}_{i=1}^{|\mathcal{M}_e|}$.
For a pair of head and tail entities $(e_h, e_t)$, document-level RE aims to predict if any relation $r \in \mathcal{R}$ exists between them, based on whether $r$ is expressed by some pair of $e_h$'s and $e_t$'s mentions in $d$.
Here, $\mathcal{C}$ and $\mathcal{R}$ are pre-defined sets of entity and relation types, respectively.
Moreover, for $(e_h, e_t)$ and each of their valid relations $r \in \mathcal{R}_{h,t}$, ER aims to identify the subset $\mathcal{V}_{h,t,r}$ of $\mathcal{S}_d$ that is sufficient to express the triplet $(e_h, e_t, r)$.

\subsection{Related Work} \label{sec:2.2}

Early research efforts on RE \cite{bach2007review, pawar2017relation} center around predicting relations for entity pairs at the sentence level \cite{zhang2017position, hendrickx2019semeval}. 
Many pattern-based \cite{califf1999relational, qu2018weakly, zhou2020nero} and neural network-based \cite{cai2016bidirectional, feng2018reinforcement, zhang2019long} models have shown impressive results. 
A recent study \cite{peng2020learning} attributes the success of these models to their ability to capture textual contexts and entity type information.

Nevertheless, since more complicated relations can only be expressed by multiple sentences, there has been a shift of focus lately towards document-level RE \cite{yao2019docred, li2016biocreative, cheng2021hacred, wu2019renet}. 
According to how an approach models contexts, there are two general trends within the domain.
Graph-based approaches \cite{nan2020reasoning, wang2020global, zeng2020double, li2020graph, zeng2021sire, xu2021discriminative, xu2021document, sahu2019inter, guo2019attention} typically infuse contexts into heuristic-based document graphs and perform multi-hop reasoning via advanced neural techniques. 
Transformer-based approaches \cite{wang2019fine, tang2020hin, huang2020entity, xu2021entity, zhou2021document, zhang2021document, xie2021eider, ye2020coreferential} leverage the strength of pre-trained language models \cite{devlin2018bert, liu2019roberta} to encode long-range contextual dependencies.
However, most prior methods only implicitly learn to capture contexts while being trained for RE.
Consequently, they experience ineffective supervision and uninterpretable model predictions.

On the contrary, we propose to explicitly teach the model to capture textual contexts and entity type information via a broad spectrum of carefully designed tasks.
Furthermore, we boost the RE performance by ensembling the results of evidence-augmented inputs.
Compared to EIDER \cite{xie2021eider}, we leverage the more precise and interpretable FER for retrieving evidence and present two different kinds of evidence-based data augmentation. 
We also save the computational cost by applying ensemble learning only to the uncertain subset of relation triplets.
As a result, our SAIS method not only enhances the RE performance due to more effective supervision, but also retrieves more accurate evidence for better interpretability.

% Besides multi-task learning, we further boosted the performance by ensembling the results of evidence-augmented inputs. 
% Although EIDER \cite{xie2021eider} adopted a similar idea, it used PER for this step and ensemble learning for all possible relation triplets. 
% As demonstrated in the qualitative analysis in Section 5.4, FER is more precise and interpretable than PER. 
% Hence, we leveraged FER instead of PER here and proposed not only pseudo document-based but also attention mask-based data augmentation. 
% Furthermore, we also aimed to preserve the computational cost and, therefore, applied ensemble learning only to the triplet subset that the model was uncertain about. 
% Again, we justified our design via the ablation study in Section 5.3.
\section{Supervising Intermediate Steps} \label{sec:3}

This section describes the tasks that explicitly supervise the model's outputs in the four intermediate steps. Together they complement the quality of RE.

\subsection{Document Encoding} \label{sec:3.1}

Given the promising performance of pre-trained language models (PLM) in various downstream tasks, we resort to PLM for encoding the document.
More specifically, for a document $d$, we insert a classifier token ``[CLS]'' and a separator token ``[SEP]'' at the start and end of each sentence $s \in \mathcal{S}_d$, respectively.
Each mention $m \in \mathcal{M}_d$ is wrapped with a pair of entity markers ``*'' \cite{zhang2017position} to indicate the position of entity mentions. 
% Then we feed the document, with alternating segment token indices for each sentence \cite{liu2019text}, into a PLM to obtain the token embeddings $\mathbf{H} \in \mathbb{R}^{N_d \times H}$ and the cross-token attention $\mathbf{A} \in \mathbb{R}^{N_d \times N_d}$.
Then we feed the document, with alternating segment token indices for each sentence \cite{liu2019text}, into a PLM:
\begin{align} \label{eq:doc_plm}
    \mathbf{H}, \mathbf{A} = \text{PLM}(d),
\end{align}
to obtain the token embeddings $\mathbf{H} \in \mathbb{R}^{N_d \times H}$ and the cross-token attention $\mathbf{A} \in \mathbb{R}^{N_d \times N_d}$.
$\mathbf{A}$ is the average of the attention heads in the last transformer layer \cite{vaswani2017attention} of the PLM. 
$N_d$ is the number of tokens in $d$, and $H$ is the embedding dimension of the PLM. 
We take the embedding of ``*" or ``[CLS]" before each mention or sentence as the corresponding mention or sentence embedding, respectively.

\subsection{Coreference Resolution (CR)} \label{sec:3.2}

As a case study, it is reported by \citet{yao2019docred} that around $17.6\%$ of relation instances in DocRED require coreference reasoning.
Hence, after encoding the document, we resolve the repeated contextual mentions to an entity via CR.
In particular, consider a pair of mentions $(m_i, m_j)$, we determine the probability of whether $m_i$ and $m_j$ refer to the same entity by passing their corresponding embeddings $\mathbf{m}_i$ and $\mathbf{m}_j$ through a group bilinear layer \cite{zheng2019learning}.
% The layer splits the embeddings into $K$ equal-sized groups ($[\mathbf{m}_i^1, \dots, \mathbf{m}_i^K] = \mathbf{m}_i$, similar for $\mathbf{m}_j$) and applies bilinear with parameter $\mathbf{W}_m^k \in \mathbb{R}^{H/K \times H/K}$ within each group: $\mathbb{P}_{i,j}^\text{CR} = \sigma \left( \sum_{k=1}^{K} \mathbf{m}_i^{k\top} \mathbf{W}_m^k \mathbf{m}_j^k + b_m \right)$,
The layer splits the embeddings into $K$ equal-sized groups ($[\mathbf{m}_i^1, \dots, \mathbf{m}_i^K] = \mathbf{m}_i$, similar for $\mathbf{m}_j$) and applies bilinear with parameter $\mathbf{W}_m^k \in \mathbb{R}^{H/K \times H/K}$ within each group:
\begin{align} \label{eq:cr_bilinear}
    \mathbb{P}_{i,j}^\text{CR} = \sigma \left( \sum_{k=1}^{K} \mathbf{m}_i^{k\top} \mathbf{W}_m^k \mathbf{m}_j^k + b_m \right),
\end{align}
where $b_m \in \mathbb{R}$ and $\sigma$ is the sigmoid function.

Since most mention pairs refer to distinct entities (each entity has only $1.34$ mentions on average in DocRED), we adopt the focal loss \cite{lin2017focal} on top of the binary cross-entropy to mitigate this extreme class imbalance: %$\ell^\text{CR}_d = - \sum_{(m_i, m_j) \in \mathcal{M}_d^2} w_{i,j}^\text{CR} \left( y_{i,j}^\text{CR} (1 - \mathbb{P}_{i,j}^\text{CR})^{\gamma^\text{CR}} \log \mathbb{P}_{i,j}^\text{CR} \right.$ $+ \left. (1-y_{i,j}^\text{CR}) (\mathbb{P}_{i,j}^\text{CR})^{\gamma^\text{CR}} \log (1-\mathbb{P}_{i,j}^\text{CR}) \right)$,
\begin{align} \label{eq:cr_loss}
    \!\! \ell^\text{CR}_d & = - \!\! \sum_{m_i \in \mathcal{M}_d} \sum_{m_j \in \mathcal{M}_d} \!\! \left( y_{i,j}^\text{CR} (1 - \mathbb{P}_{i,j}^\text{CR})^{\gamma^\text{CR}} \log \mathbb{P}_{i,j}^\text{CR} \!\! \right. \nonumber \\ & \left. \!\!\!\!\!\!\! + \, (1-y_{i,j}^\text{CR}) (\mathbb{P}_{i,j}^\text{CR})^{\gamma^\text{CR}} \log (1-\mathbb{P}_{i,j}^\text{CR}) \right) w_{i,j}^\text{CR}\, ,
\end{align}
where $y_{i,j}^\text{CR} = 1$ if $m_i$ and $m_j$ refer to the same entity, and $0$ otherwise. 
Class weight $w_{i,j}^\text{CR}$ is inversely proportional to the frequency of $y_{i,j}^\text{CR}$, and $\gamma^\text{CR}$ is a hyperparameter.

\subsection{Entity Typing (ET)} \label{sec:3.3}

In a pair of entities, the type information can be used to filter out impossible relations.
Therefore, we regularize entity embeddings via ET.
More specifically, we first derive the embedding of an entity $e$ by integrating the embeddings of its mentions $\mathcal{M}_e$ via logsumexp pooling \cite{jia2019document}: $\mathbf{e} = \log \sum_{m \in \mathcal{M}_e} \exp (\mathbf{m})$. 
Since entity $e$ could appear either at the head or tail in an entity pair, we distinguish between the head entity embedding $\mathbf{e}'_h$ and the tail entity embedding  $\mathbf{e}'_t$ via two separate linear layers: %$\mathbf{e}'_{h} = \mathbf{W}_{e_h} \mathbf{e} + \mathbf{b}_{e_h}$, $\mathbf{e}'_{t} = \mathbf{W}_{e_t} \mathbf{e} + \mathbf{b}_{e_t}$,
\begin{align} \label{eq:et_ht_embs}
    \mathbf{e}'_{h} = \mathbf{W}_{e_h} \mathbf{e} + \mathbf{b}_{e_h}, \;\!\; \mathbf{e}'_{t} = \mathbf{W}_{e_t} \mathbf{e} + \mathbf{b}_{e_t},
\end{align}
where $\mathbf{W}_{e_h}, \mathbf{W}_{e_t} \in \mathbb{R}^{H \times H}$ and $\mathbf{b}_{e_h}, \mathbf{b}_{e_t} \in \mathbb{R}^{H}$.

However, no matter where $e$ appears in an entity pair, its head and tail embeddings should always preserve $e$'s type information. 
Hence, we calculate the probability of which entity type $e$ belongs to by passing $\mathbf{e}'_\nu$ for $\nu \in \{h,t\}$ through a linear layer %$\mathbb{P}_{e}^{\text{ET}} = \varsigma (\mathbf{W}_c  \tanh(\mathbf{e}'_\nu) + \mathbf{b}_c)$, 
\begin{align} \label{eq:et_linear}
    \mathbb{P}_{e}^{\text{ET}} & = \varsigma (\mathbf{W}_c  \tanh(\mathbf{e}'_\nu) + \mathbf{b}_c)\, , 
\end{align}
followed by the multi-class cross-entropy loss: %$\ell^{\text{ET}}_d = - \sum_{e \in \mathcal{E}_d} \sum_{c \in \mathcal{C}} y_{e,c}^{\text{ET}} \log \mathbb{P}_{e,c}^{\text{ET}}$,
\begin{align} \label{eq:et_loss}
    \ell^{\text{ET}}_d & = - \sum_{e \in \mathcal{E}_d} \sum_{c \in \mathcal{C}} y_{e,c}^{\text{ET}} \log \mathbb{P}_{e,c}^{\text{ET}}\, , 
\end{align}
where $\mathbf{W}_c \in \mathbb{R}^{|\mathcal{C}| \times H}$, $\mathbf{b}_c \in \mathbb{R}^{|\mathcal{C}|}$, and $\varsigma$ is the softmax function. $y_{e,c}^\text{ET} = 1$ if $e$ is of entity type $c$, and $0$ otherwise.

\subsection{Pooled Evidence Retrieval (PER)} \label{sec:3.4}

To further capture textual contexts, we explicitly guide the attention in the PLM to the supporting sentences of each entity pair via PER.
That is, we want to identify the pooled evidence set $\mathcal{V}_{h,t} = \cup_{r \in \mathcal{R}_{h,t}} \mathcal{V}_{h,t,r}$ in $d$ that is important to an entity pair $(e_h, e_t)$, regardless of the specific relation expressed by a particular sentence $s \in \mathcal{V}_{h,t}$. 
% In this case, given $(e_h, e_t)$, we first compute a unique context embedding $\mathbf{c}_{h,t} = \mathbf{H}^\top \frac{ \mathbf{A}_h \otimes \mathbf{A}_t }{\mathbf{1}^{\top} ( \mathbf{A}_h \otimes \mathbf{A}_t)}$.
In this case, given $(e_h, e_t)$, we first compute a unique context embedding $\mathbf{c}_{h,t}$ based on the cross-token attention from Equation~\ref{eq:doc_plm}:
\begin{align} \label{eq:per_context}
    \mathbf{c}_{h,t} = \mathbf{H}^\top \frac{ \mathbf{A}_h \otimes \mathbf{A}_t }{\mathbf{1}^{\top} ( \mathbf{A}_h \otimes \mathbf{A}_t)}\, .
\end{align}
Here, $\otimes$ is the element-wise product.
$\mathbf{A}_h$ is $e_h$'s attention to all the tokens in the document (i.e., the average of $e_h$'s mention-level attention).  
Similar for $\mathbf{A}_t$.
Then we measure the probability of whether a sentence $s \in \mathcal{S}_d$ is part of the pooled supporting evidence $\mathcal{V}_{h,t}$ by passing 
$(e_h, e_t)$'s context embedding $\mathbf{c}_{h,t}$ and sentence $s$' embedding $\mathbf{s}$ through a group bilinear layer: %$\mathbb{P}^\text{PER}_{h,t,s} = \sigma \left( \sum_{k=1}^{K} \mathbf{c}_{h,t}^{k\top}  \mathbf{W}_{p}^{k}\mathbf{s}^{k}  + b_p \right)$,
\begin{align} \label{eq:per_bilinear}
    \mathbb{P}^\text{PER}_{h,t,s} = \sigma \left( \sum_{k=1}^{K} \mathbf{c}_{h,t}^{k\top}  \mathbf{W}_{p}^{k}\mathbf{s}^{k}  + b_p \right),
\end{align}
where $\mathbf{W}_p^k \in \mathbb{R}^{H/K \times H/K}$ and $b_p \in \mathbb{R}$.

Again, we face a severe class imbalance here, since most entity pairs ($97.1\%$ in DocRED) do not have valid relations or supporting evidence. 
As a result, similar to Section~\ref{sec:3.2}, we also use the focal loss with the binary cross-entropy: %$\ell^\text{PER}_d = - \sum_{(e_i, e_j) \in \mathcal{E}_d^2} \sum_{s \in \mathcal{S}_d} w_{h,t,s}^\text{PER} \left( y_{h,t,s}^\text{PER} (1 - \mathbb{P}_{h,t,s}^\text{PER})^{\gamma^\text{PER}} \right. $ $ \left. \log \mathbb{P}_{h,t,s}^\text{PER} \!+\! (1\!-\!y_{h,t,s}^\text{PER}) (\mathbb{P}_{h,t,s}^\text{PER})^{\gamma^\text{PER}} \log (1-\mathbb{P}_{h,t,s}^\text{PER}) \right)$,
\begin{align} \label{eq:per_loss}
    \ell^\text{PER}_d = & - \sum_{e_h \in \mathcal{E}_d} \sum_{e_t \in \mathcal{E}_d} \sum_{s \in \mathcal{S}_d} \left( y_{h,t,s}^\text{PER} (1 - \mathbb{P}_{h,t,s}^\text{PER})^{\gamma^\text{PER}} \right. \nonumber \\ & \log \mathbb{P}_{h,t,s}^\text{PER} + \, (1-y_{h,t,s}^\text{PER}) (\mathbb{P}_{h,t,s}^\text{PER})^{\gamma^\text{PER}} \nonumber \\ & \left. \log (1-\mathbb{P}_{h,t,s}^\text{PER}) \right) w_{h,t,s}^\text{PER}\, ,
\end{align}
where $y_{h,t,s}^\text{PER} = \mathbbm{1} \{ s \in \mathcal{V}_{h,t} \}$, class weight $w_{h,t,s}^\text{PER}$ is inversely proportional to the frequency of $y_{h,t,s}^\text{PER}$, and $\gamma^\text{PER}$ is a hyperparameter.

\subsection{Fine-grained Evidence Retrieval (FER)} \label{sec:3.5}

In addition to PER, we would like to further refine $\mathcal{V}_{h,t}$, since an entity pair could have multiple valid relations and, correspondingly, multiple sets of evidence. 
As a result, we explicitly train the model to recover contextual evidence unique to a triplet $(e_h, e_t, r)$ via FER for better interpretability.
More specifically, given $(e_h, e_t, r)$, we first generate a triplet embedding $\mathbf{l}_{h,t,r}$ by merging $\mathbf{e}_h$, $\mathbf{e}_t$, $\mathbf{c}_{h,t}$, and $r$'s relation embedding $\mathbf{r}$ via a linear layer: %$\mathbf{l}_{h,t,r} = \tanh( \mathbf{W}_l [\mathbf{e}_h\|\mathbf{e}_t\|\mathbf{c}_{h,t}\|\mathbf{r}] + \mathbf{b}_{l})$,
\begin{align} \label{eq:fer_triplet_emb}
    \mathbf{l}_{h,t,r} = \tanh( \mathbf{W}_l [\mathbf{e}_h\|\mathbf{e}_t\|\mathbf{c}_{h,t}\|\mathbf{r}] + \mathbf{b}_{l})\, ,
\end{align}
where $\mathbf{W}_l \in \mathbb{R}^{H \times 4H}$, $\mathbf{b}_l \in \mathbb{R}^{H}$, $\|$ represents concatenation, and $\mathbf{r}$ is initialized from the embedding matrix of the PLM.

Similarly, we use a group bilinear layer to assess the probability of whether a sentence $s \in \mathcal{S}_d$ is included in the fine-grained evidence set $\mathcal{V}_{h,t,r}$: % $\mathbb{P}^\text{FER}_{h,t,r,s} = \sigma \left( \sum_{k=1}^{K} \mathbf{l}_{h,t,r}^{k\top}  \mathbf{W}_{f}^{k}\mathbf{s}^{k}  + b_f \right)$,
\begin{align} \label{eq:fer_bilinear}
    \mathbb{P}^\text{FER}_{h,t,r,s} = \sigma \left( \sum_{k=1}^{K} \mathbf{l}_{h,t,r}^{k\top}  \mathbf{W}_{f}^{k}\mathbf{s}^{k}  + b_f \right),
\end{align}
where $\mathbf{W}_f^k \in \mathbb{R}^{H/K \times H/K}$ and $b_f \in \mathbb{R}$.

Since FER only involves entity pairs with valid relations, the class imbalance is milder here than in PER.
Hence, let $y_{h,t,r,s}^\text{FER} = \mathbbm{1} \{ s \in \mathcal{V}_{h,t,r} \}$,
we deploy the standard binary cross-entropy loss:
% $\ell^\text{FER}_d =  - \sum_{(e_i, e_j) \in \mathcal{E}_d^2} \sum_{r \in \mathcal{R}_{h,t}} \sum_{s \in \mathcal{S}_d} \left( y_{h,t,r,s}^\text{FER} \log \mathbb{P}_{h,t,r,s}^\text{FER} \right.$ $ \left.+ (1-y_{h,t,r,s}^\text{FER}) \log (1-\mathbb{P}_{h,t,r,s}^\text{FER}) \right)$.
% where $y_{h,t,r,s}^\text{FER} = \mathbbm{1} \{ s \in \mathcal{V}_{h,t,r} \}$.
\begin{align} \label{eq:fer_loss}
    \! \ell^\text{FER}_d \!= & - \!\!\! \sum_{e_i \in \mathcal{E}_d} \sum_{e_j \in \mathcal{E}_d} \sum_{r \in \mathcal{R}_{h,t}} \sum_{s \in \mathcal{S}_d} \left( y_{h,t,r,s}^\text{FER} \log \mathbb{P}_{h,t,r,s}^\text{FER} \!\! \right. \nonumber \\ & \left. + \, (1-y_{h,t,r,s}^\text{FER}) \log (1-\mathbb{P}_{h,t,r,s}^\text{FER}) \right) .
\end{align}

\subsection{Relation Extraction (RE)} \label{sec:3.6}

Based on the four complementary tasks introduced above, for an entity pair $(e_h, e_t)$, we encode relevant contexts in $\mathbf{c}_{h,t}$ and preserve entity type information in $\mathbf{e}'_h$ and $\mathbf{e}'_t$. 
% Ultimately, we acquire the contexts needed by the head and tail entity from $\mathbf{c}_{h,t}$ via two separate linear layers, and then combine them with the type information to generate the head entity representation $\mathbf{e}''_{h} = \tanh \left( (\mathbf{W}_{c_h} \mathbf{c}_{h,t} + \mathbf{b}_{c_h}) + \mathbf{e}'_{h} \right)$ and the tail entity representation $\mathbf{e}''_{t} = \tanh \left( (\mathbf{W}_{c_t} \mathbf{c}_{h,t} + \mathbf{b}_{c_t}) + \mathbf{e}'_{t} \right)$,
% where $\mathbf{W}_{c_h}, \mathbf{W}_{c_t} \in \mathbb{R}^{H \times H}$ and $\mathbf{b}_{c_h}, \mathbf{b}_{c_t} \in \mathbb{R}^{H}$.
Ultimately, we acquire the contexts needed by the head and tail entities from $\mathbf{c}_{h,t}$ via two separate linear layers:
\begin{align} \label{eq:re_ht_contexts}
    \! \mathbf{c}'_{h} = \mathbf{W}_{c_h} \mathbf{c}_{h,t} + \mathbf{b}_{c_h},  \mathbf{c}'_{t} = \mathbf{W}_{c_t} \mathbf{c}_{h,t} + \mathbf{b}_{c_t},
\end{align}
where $\mathbf{W}_{c_h}, \mathbf{W}_{c_t} \in \mathbb{R}^{H \times H}$ and $\mathbf{b}_{c_h}, \mathbf{b}_{c_t} \in \mathbb{R}^{H}$, and then combine them with the type information to generate the head and tail entity representations:
\begin{align} \label{eq:re_ht_reps}
    \!\mathbf{e}''_{h} = \tanh( \mathbf{e}'_{h} + \mathbf{c}'_h),\; \mathbf{e}''_{t} = \tanh( \mathbf{e}'_{t} + \mathbf{c}'_t).
\end{align}

Next, a group bilinear layer is utilized to calculate the logit of how likely a relation $r \in \mathcal{R}$ exists between $e_h$ and $e_t$: %$\mathbb{L}^\text{RE}_{h,t,r} = \sum_{k=1}^{K} \mathbf{e}_{h}''^{k\top}  \mathbf{W}_{r}^{k} \mathbf{e}_{t}''^{k}  + b_r$,
\begin{align} \begin{aligned} \label{eq:re_ori_logit}
    \mathbb{L}^\text{RE}_{h,t,r} = \sum_{k=1}^{K} \mathbf{e}_{h}''^{k\top}  \mathbf{W}_{r}^{k} \mathbf{e}_{t}''^{k}  + b_r\, ,
\end{aligned} \end{align}
where $\mathbf{W}_r^k \in \mathbb{R}^{H/K \times H/K}$ and $b_r \in \mathbb{R}$.

As discussed earlier, only a small portion of entity pairs have valid relations, among which multiple relations could co-exist between a pair.
Therefore, to deal with the problem of multi-label imbalanced classification, we follow \citet{zhou2021document} by introducing a threshold relation class $\text{TH}$ and adopting an adaptive threshold loss: %$\ell^{\text{RE}}_d = - \sum_{(e_h, e_t) \in \mathcal{E}_d^2} \left( \sum_{r \in \mathcal{P}_{h,t}} \log \frac{\exp \mathbb{L}^\text{RE}_{h,t,r}}{\sum_{r'\in\mathcal{P}_{h,t}\cup\{\text{TH}\}} \mathbb{L}^\text{RE}_{h,t,r'}} \right.$ $\left. + \log \frac{\exp \mathbb{L}^\text{RE}_{h,t,\text{TH}}}{\sum_{r'\in\mathcal{N}_{h,t}\cup\{\text{TH}\}} \mathbb{L}^\text{RE}_{h,t,r'}} \right)$
% \begin{align} \label{eq:re_loss}
%     \ell^{\text{RE}}_d = & \sum_{e_h \in \mathcal{E}_d} \sum_{e_t \in \mathcal{E}_d} \ell^{\text{RE}}_{h,t} \;, \;\;\;\; \text{where} \;\; \nonumber
%     \\ \ell^{\text{RE}}_{h,t} = & - \left[ \sum_{r \in \mathcal{P}_{h,t}} \log \left( \frac{\exp \mathbb{L}^\text{RE}_{h,t,r}}{\sum_{r'\in\mathcal{P}_{h,t}\cup\{\text{TH}\}} \mathbb{L}^\text{RE}_{h,t,r'}} \right) \right. \nonumber
%     \\ & \left. + \log \left( \frac{\exp \mathbb{L}^\text{RE}_{h,t,\text{TH}}}{\sum_{r'\in\mathcal{N}_{h,t}\cup\{\text{TH}\}} \mathbb{L}^\text{RE}_{h,t,r'}} \right) \right].
% \end{align}
\begin{align} \label{eq:re_loss}
    \ell^{\text{RE}}_d = & - \sum_{e_h \in \mathcal{E}_d} \sum_{e_t \in \mathcal{E}_d} \nonumber \\ 
    & \left[ \sum_{r \in \mathcal{P}_{h,t}} \log \left( \frac{\exp \mathbb{L}^\text{RE}_{h,t,r}}{\sum_{r'\in\mathcal{P}_{h,t}\cup\{\text{TH}\}} \mathbb{L}^\text{RE}_{h,t,r'}} \right) \right. \nonumber
    \\ & \left. + \log \left( \frac{\exp \mathbb{L}^\text{RE}_{h,t,\text{TH}}}{\sum_{r'\in\mathcal{N}_{h,t}\cup\{\text{TH}\}} \mathbb{L}^\text{RE}_{h,t,r'}} \right) \right].
\end{align}
In essence, we aim to increase the logits of valid relations $\mathcal{P}_{h,t}$ and decrease the logits of invalid relations $\mathcal{N}_{h,t}$, both relative to $\text{TH}$.

Overall, with the goal of improving the model's RE performance by better capturing entity type information and textual contexts, we have designed four tasks to explicitly supervise the model's outputs in the corresponding intermediate steps.
To this end, we visualize the entire pipeline $\text{SAIS}_\text{All}^\text{O}$ in Appendix~\ref{ap:a} and integrate all the tasks by minimizing the multi-task learning objective %$\ell = \sum_{d \in \mathcal{D}_\text{train}} \!\left( \ell^\text{RE}_d \!+\! \sum_{\text{Task} \in \{ \text{CR},\, \text{ET},\, \text{PER},\, \text{FER} \}}\! \eta^\text{Task} \ell^\text{Task}_d \!\right)$,
\begin{align} \label{eq:overall_loss}
    \ell = \sum_{d \in \mathcal{D}_\text{train}} \left( \ell^\text{RE}_d + \sum_{\text{Task}} \eta^\text{Task} \ell^\text{Task}_d \right) ,
\end{align}
where $\text{Task} \in \{\text{CR, ET, PER, FER}\}$. 
$\eta^\text{Task}$'s are hyperparameters balancing the relative task weight.

During inference with the current pipeline $\text{SAIS}_\text{All}^\text{O}$, we predict if a triplet $(e_h, e_t, r)$ is valid (i.e., if relation $r$ exists between entity pair $(e_h, e_t)$) by checking if its logit is larger than the corresponding threshold logit (i.e., $\mathbb{L}_{h,t,r}^\text{RE} > \mathbb{L}_{h,t,\text{TH}}^\text{RE}$).
For each predicted triplet $(e_h, e_t, r)$, we assess if a sentence $s$ belongs to the evidence set $\mathcal{V}_{h,t,r}$ by checking if $\mathbb{P}_{h,t,r,s}^\text{FER} > \alpha^\text{FER}$ where $\alpha^\text{FER}$ is a threshold.
\section{Augmenting Intermediate Steps} \label{sec:4}

We further improve RE after training the pipeline $\text{SAIS}_\text{All}^\text{O}$ by augmenting the intermediate steps in $\text{SAIS}_\text{All}^\text{O}$ with the retrieved evidence from FER.

\subsection{When to Augment Intermediate Steps} \label{sec:4.1}

The evidence predicted by FER is unique to each triplet $(e_h, e_t, r)$.
However, consider the total number of all possible triplets (around 40 million in the develop set of DocRED), it is computationally prohibitive to augment the inference result of each triplet with individually predicted evidence.
Instead, following the idea of selective prediction \cite{el2010foundations}, we identify the triplet subset $\mathcal{U}$ for which the model is \textit{uncertain} about its relation predictions with the original pipeline $\text{SAIS}_\text{All}^\text{O}$.
More specifically, we set the model's confidence for $(e_h, e_t, r)$ as $\mathbb{L}_{h,t,r}^\text{O} = \mathbb{L}_{h,t,r}^\text{RE} - \mathbb{L}_{h,t,\text{TH}}^\text{RE}$.
Then, the uncertain set $\mathcal{U}$ consists of triplets with the lowest $\theta\%$ absolute confidence $|\mathbb{L}_{h,t,r}^\text{O}|$.
Consequently, we reject the original relation predictions for $(e_h, e_t, r) \in \mathcal{U}$ and apply evidence-based data augmentation to enhance the performance (more details in Section~\ref{sec:4.2}).

\begin{figure}[t!]
\centering
\includegraphics[width=0.85\columnwidth]{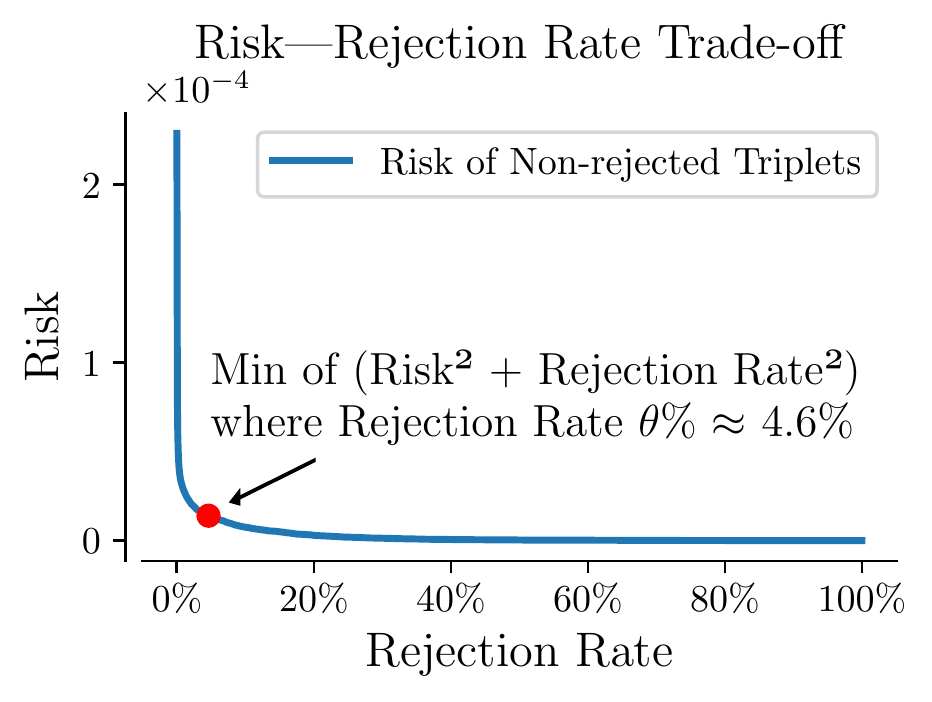} % 0.655
\caption{Trade-off between risk and rejection rate on the develop set of DocRED.}
\label{fig:2}
\end{figure}

To determine the rejection rate $\theta\%$ (note that $\theta\%$ is NOT a hyperparameter), we first sort all the triplets in the develop set based on their absolute confidence $|\mathbb{L}_{h,t,r}^\text{O}|$.
When $\theta\%$ increases, the risk (i.e., inaccuracy rate) of the remaining triplets that are not in $\mathcal{U}$ is expected to decrease, and vice versa.
On the one hand, we wish to reduce the risk for more accurate relation predictions; on the other hand, we want a low rejection rate so that data augmentation on a small rejected set incurs little computational cost.
To balance this trade-off, we set $\theta\%$ as the rate that achieves the minimum of $\text{risk}^2 + \text{rejection rate}^2$. 
As shown in Figure~\ref{fig:2}, we find $\theta\% \approx 4.6\%$ in the develop set of DocRED.
In practice, we can further limit the maximum number of rejected triplets per entity pair.
By setting it as $10$ in experiments, we reduce the size of $\mathcal{U}$ to only $1.5\%$ of all the triplets in the DocRED develop set.

\begin{table*}[t!]
  \centering
  \setlength\tabcolsep{3.8pt}
\resizebox{\textwidth}{!}{
  \begin{tabular}{lcccccc} 
    \toprule
         & \multicolumn{3}{c}{\textbf{DocRED Dev}} & \multicolumn{3}{c}{\textbf{DocRED Test}} \\
    \cmidrule(lr){2-4} \cmidrule(lr){5-7}
        \textbf{Model} & \multicolumn{2}{c}{Relation} & Evidence & \multicolumn{2}{c}{Relation} & Evidence \\
    \cmidrule(lr){2-3} \cmidrule(lr){4-4} \cmidrule(lr){5-6} \cmidrule(lr){7-7}
        & Ign F1 & F1 & F1 & Ign F1 & F1 & F1 \\
    \midrule
    \midrule
        % \text{CNN}\text{CNN} \cite{yao2019docred} & 41.58 & 43.45 & - & 40.33 & 42.26 & - \\
        % $\text{GAT}$ \cite{velivckovic2017graph} & 45.17 & 51.44 & - & 47.36 & 49.51 & - \\
        % $\text{BiLSTM}$ \cite{yao2019docred} & 48.87 & 50.94 & 44.07 & 48.78 & 51.06 & 43.83 \\
        % $\text{GCNN}$ \cite{sahu2019inter} & 46.22 & 51.52 & - & 49.59 & 51.62 & - \\
        % $\text{EoG}$ \cite{christopoulou2019connecting} & 45.94 & 52.15 & - & 49.48 & 51.82 & - \\
        % $\text{AGGCN}$ \cite{guo2019attention} & 46.29 & 52.47 & - & 48.89 & 51.45 & - \\
    % \midrule
        % $\text{GEDA-BERT}_\text{base}$ \cite{li2020graph} & 54.52 & 56.16 & - & 53.71 & 55.74 & - \\
        % $\text{GLRE-BERT}_\text{base}$ \cite{wang2020global} & - & - & - & 55.40 & 57.40 & - \\
        % $\text{LSR-BERT}_\text{base}$ \cite{nan2020reasoning} & 52.43 & 59.00 & - & 56.97 & 59.05 & - \\
        $\text{HeterGSAN-BERT}_\text{base}$ \cite{xu2021document} & 58.13 & 60.18 & - & 57.12 & 59.45 & - \\
        $\text{GAIN-BERT}_\text{base}$ \cite{zeng2020double} & 59.14 & 61.22 & - & 59.00 & 61.24 & - \\
        $\text{DRN-BERT}_\text{base}$ \cite{xu2021discriminative} & 59.33 & 61.39 & - & 59.15 & 61.37 & - \\
        $\text{SIRE-BERT}_\text{base}$ \cite{zeng2021sire} & 59.82 & 61.60 & - & 60.18 & 62.05 & - \\ 
    \midrule
        $\text{BERT}_\text{base}$ \cite{wang2019fine} & - & 54.16 & - & - & 53.20 & - \\
        % $\text{BERT-TS}_\text{base}$ \cite{wang2019fine} & - & 54.42 & - & - & 53.92 & - \\
        % $\text{HIN-BERT}_\text{base}$ \cite{tang2020hin} & 54.29 & 56.31 & - & 53.70 & 55.60 & - \\
        % $\text{CorefBERT}_\text{base}$ \cite{ye2020coreferential} & 55.32 & 57.51 & - & 54.54 & 56.96 & - \\
        $\text{E2GRE-BERT}_\text{base}$ \cite{huang2020entity} & 55.22 & 58.72 & 47.14 & - & - & - \\
        $\text{SSAN-BERT}_\text{base}$ \cite{xu2021entity} & 57.03 & 59.19 & - & 56.06 & 58.41 & - \\
        $\text{ATLOP-BERT}_\text{base}$ \cite{zhou2021document} & 59.22 & 61.09 & - & 59.31 & 61.30 & - \\
        $\text{DocuNet-BERT}_\text{base}$ \cite{zhang2021document} & 59.86 & 61.83 & - & 59.93 & 61.86 & - \\
        $\text{Eider-BERT}_\text{base}$ \cite{xie2021eider} & \textbf{60.51} & 62.48 & 50.71 & 60.42 & 62.47 & 51.27 \\
    \midrule
        $\text{SAIS}^\text{B}_\text{All}\text{-BERT}_\text{base}$ (Ours) & 59.98 $\pm$ 0.13 & \textbf{62.96} $\pm$ \textbf{0.11} & \textbf{53.70} $\pm$ \textbf{0.21} & \textbf{60.96} & \textbf{62.77} & \textbf{52.88} \\
    \midrule
    \midrule
        % $\text{BERT}_\text{large}$ \cite{ye2020coreferential} & 56.51 & 58.70 & - & 56.01 & 58.31 & - \\
        % $\text{CorefBERT}_\text{large}$ \cite{ye2020coreferential} & 56.82 & 59.01 & - & 56.40 & 58.83 & - \\
        $\text{RoBERTa}_\text{large}$ \cite{ye2020coreferential} & 57.19 & 59.40 & - & 57.74 & 60.06 & - \\
        % $\text{CorefRoBERTa}_\text{large}$ \cite{ye2020coreferential} & 57.35 & 59.43 & - & 57.90 & 60.25 & - \\
        $\text{SSAN-RoBERTa}_\text{large}$ \cite{xu2021entity} & 60.25 & 62.08 & - & 59.47 & 61.42 & - \\
        $\text{E2GRE-RoBERTa}_\text{large}$ \cite{huang2020entity} & - & - & - & 60.30 & 62.50 & 50.50 \\
        $\text{ATLOP-RoBERTa}_\text{large}$ \cite{zhou2021document} & 61.32 & 63.18 & - & 61.39 & 63.40 & - \\
        $\text{DocuNet-RoBERTa}_\text{large}$ \cite{zhang2021document} & 62.23 & 64.12 & - & 62.39 & 64.55 & - \\
        $\text{Eider-RoBERTa}_\text{large}$ \cite{xie2021eider} & \textbf{62.34} & 64.27 & 52.54 & 62.85 & 64.79 & 53.01 \\
    \midrule
        $\text{SAIS}^\text{B}_\text{All}\text{-RoBERTa}_\text{large}$ (Ours) & 62.23 $\pm$ 0.15 & \textbf{65.17} $\pm$ \textbf{0.08} & \textbf{55.84} $\pm$ \textbf{0.23} & \textbf{63.44} & \textbf{65.11} & \textbf{55.67} \\
    \bottomrule
  \end{tabular}
}
  \caption{RE and ER results ($\%$) on DocRED. 
  Ign F1 refers to the F1 score excluding the relation instances mentioned in the train set.
  Baselines using $\text{BERT}_\text{base}$ are separated into the graph-based (upper) and transformer-based (lower) groups. 
  We report the test scores from the official scoreboard and the baseline scores from the corresponding papers.
  $\text{SAIS}_\text{All}^\text{B}$ achieves state-of-the-art performance on both RE and ER.
%   Baselines delivering less competitive results are omitted here.
  Full details in Appendix~\ref{ap:c}.}
  \label{tab:docred_result}
\end{table*}

\subsection{How to Augment Intermediate Steps} \label{sec:4.2}

Consider a triplet $(e_h, e_t, r) \in \mathcal{U}$. We first assume its validity and calculate the probability $\mathbb{P}_{h,t,r,s}^\text{FER}$ of a sentence $s$ being part of $\mathcal{V}_{h,t,r}$ based on Section~\ref{sec:3.5}.
Then in a similar way to how $\mathbb{L}_{h,t,r}^\text{O}$ is generated with $\text{SAIS}_\text{All}^\text{O}$, we design two types of evidence-based data augmentation as follows:
\begin{description}[style=unboxed,nolistsep]
    \item [Pseudo Document-based ($\text{SAIS}_\text{All}^\text{D}$):] 
    Construct a pseudo document using sentences with $\mathbb{P}_{h,t,r,s}^\text{FER} > \alpha^\text{FER}$ and feed it into the original pipeline to get the confidence $\mathbb{L}_{h,t,r}^\text{D}$.
    \item [Attention Mask-based ($\text{SAIS}_\text{All}^\text{M}$):] 
    Formulate a mask $\mathbf{P}_{h,t,r}^\text{FER} \in \mathbb{R}^{N_d}$ based on $\mathbb{P}_{h,t,r,s}^\text{FER}$ and modify the context embedding to $\mathbf{c}_{h,t} = \mathbf{H}^\top \frac{ \mathbf{A}_h \otimes \mathbf{A}_t \otimes \mathbf{P}_{h,t,r}^\text{FER} }{\mathbf{1}^{\top} ( \mathbf{A}_h \otimes \mathbf{A}_t \otimes \mathbf{P}_{h,t,r}^\text{FER})}$. Maintain the rest of the pipeline and get the confidence $\mathbb{L}_{h,t,r}^\text{M}$.
\end{description}
Following \citet{xie2021eider}, we ensemble $\mathbb{L}_{h,t,r}^\text{D}$, $\mathbb{L}_{h,t,r}^\text{M}$, and the original confidence $\mathbb{L}_{h,t,r}^\text{O}$ with a blending parameter $\tau_r \in \mathbb{R}$ \cite{wolpert1992stacked} for each relation $r \in \mathcal{R}$ as 
\begin{align} \label{eq:blend}
    \mathbb{P}_{h,t,r}^{\text{B}} & = \sigma(\mathbb{L}_{h,t,r}^\text{B}) \nonumber \\ & = \sigma (\mathbb{L}_{h,t,r}^\text{O} + \mathbb{L}_{h,t,r}^\text{D} + \mathbb{L}_{h,t,r}^\text{M} - \tau_r).
\end{align}
These parameters are trained by minimizing the binary cross-entropy loss on $\mathcal{U}$ of the develop set: 
\begin{align} \label{eq:blend_loss}
    \ell^{B} = - & \sum_{(e_h, \,e_t, \,r) \,\in\, \mathcal{U}} \left( y^\text{RE}_{h,t,r} \log \mathbb{P}^\text{B}_{h,t,r} \right. \nonumber \\ & \left. +\; (1 \!-\! y^\text{RE}_{h,t,r}) \log (1-\mathbb{P}^\text{B}_{h,t,r}) \right),
\end{align}
where $y^\text{RE}_{h,t,r} = 1$ if $(e_h, e_t, r)$ is valid, and $0$ otherwise. 
When making relation predictions for $(e_h, e_t, r) \in \mathcal{U}$, we check whether its blended confidence is positive (i.e., $\mathbb{L}_{h,t,r}^\text{B} > 0$).

In this way, we improve the RE performance when the model is uncertain about its original predictions and save the computational cost when the model is confident.
The overall steps for evidence-based data augmentation and ensemble inference $\text{SAIS}_\text{All}^\text{B}$ are summarized in Appendix~\ref{ap:b}.
These steps are executed only after the training of $\text{SAIS}_\text{All}^\text{O}$ and, therefore, adds negligible computational cost.
\section{Experiments} \label{sec:5}

\subsection{Experiment Setup} \label{sec:5.1}

% \paragraph{Datasets}
We evaluate the proposed SAIS method on the following three document-level RE benchmarks.
DocRED \cite{yao2019docred} is a large-scale crowd-sourced dataset based on Wikipedia articles.
It consists of $97$ relation types, seven entity types, and $\numprint{5053}$ documents in total, where each document has $19.5$ entities on average. 
CDR \cite{li2016biocreative} and GDA \cite{wu2019renet} are two biomedical datasets where CDR studies the binary interactions between disease and chemical concepts with $\numprint{1500}$ documents and GDA studies the binary relationships between gene and disease with $\numprint{30192}$ documents.
We follow \citet{christopoulou2019connecting} for splitting the train and develop sets.

% \paragraph{Implementation}
We run our experiments on one Tesla A6000 GPU and carry out five trials with different seeds to report the mean and one standard error.
Based on Huggingface \cite{wolf2019huggingface}, we apply cased BERT-base \cite{devlin2018bert} and RoBERTa-large \cite{liu2019roberta} for DocRED and cased SciBERT \cite{beltagy2019scibert} for CDR and GDA.
The embedding dimension $H$ of BERT or SciBERT is $768$, and that of RoBERTa is $\numprint{1024}$.
The number of groups $K$ in all group bilinear layers is $64$.

% \paragraph{Hyperparameters}
For the general hyperparameters of language models, we follow the setting in \cite{zhou2021document}.
The learning rate for fine-tuning BERT is $5\mathrm{e}{-5}$, that for fine-tuning RoBERTa or SciBERT is $2\mathrm{e}{-5}$, and that for training the other parameters is $1\mathrm{e}{-4}$.
All the trials are optimized by AdamW \cite{loshchilov2017decoupled} for $20$ epochs with early stopping and a linearly decaying scheduler \cite{goyal2017accurate} whose warm-up ratio $=6\%$.
Each batch contains $4$ documents and the gradients of model parameters are clipped to a maximum norm of $1$.

For the unique hyperparameters of our method, we choose $2$ from $\{1, 1.5, 2\}$ for the focal hyperparameters $\gamma^\text{CR}$ and $\gamma^\text{PER}$ based on the develop set.
We also follow \citet{xie2021eider} for setting the FER prediction threshold $\alpha^\text{FER}$ as $0.5$ and all the relative task weights $\eta^\text{Task}$ for $\text{Task} \in \{\text{CR}, \text{ET}, \text{PER}, \text{FER}\}$ as $0.1$.

\subsection{Quantitative Evaluation} \label{sec:5.2}

Besides RE, DocRED also suggests to predict the supporting evidence for each relation instance.
Therefore, we apply $\text{SAIS}_\text{All}^\text{B}$ to both RE and ER.
We report the results of $\text{SAIS}_\text{All}^\text{B}$ as well as existing graph-based and transformer-based baselines in Table~\ref{tab:docred_result}\footnote{For a fair comparison, we report the scores of SSAN \cite{xu2021entity} without being pretrained on an extra dataset.} (full details in Appendix~\ref{ap:c}).
Generally, thanks to PLMs' strength in modeling long-range dependencies, transformer-based methods perform better on RE than graph-based methods.
Moreover, most earlier approaches are not capable of ER despite the interpretability ER adds to the predictions.
In contrast, our $\text{SAIS}_\text{All}^\text{B}$ method not only establishes a new state-of-the-art result on RE, but also outperforms the runner-up significantly on ER. 

Since neither CDR nor GDA annotates evidence sentences, we apply $\text{SAIS}_\text{RE+CR+ET}^\text{O}$ here.
It is trained with RE, CR, and ET and infers without data augmentation.
As shown in Table~\ref{tab:bio_result} (full details in Appendix~\ref{ap:c}), our method improves the prior best RE F1 scores by $2.7\%$ and $1.8\%$ absolutely on CDR and GDA, respectively.
It indicates that our proposed method can still improve upon the baselines even if only part of the four complementary tasks are annotated and operational.

\subsection{Ablation Study} \label{sec:5.3}

To investigate the effectiveness of each of the four complementary tasks proposed in Section~\ref{sec:3}, we carry out an extensive ablation study on the DocRED develop set by training SAIS with all possible combinations of those tasks.
As shown in Table~\ref{tab:ablation_task}, without any complementary tasks, the RE performance of SAIS is comparable to ATLOP \cite{zhou2021document} due to similar neural architectures.
When only one complementary task is allowed, PER is the most effective single task, followed by ET.
Although FER is functionally analogous to PER, since FER only involves the small subset of entity pairs with valid relations, the performance gain brought by FER alone is limited.
When two tasks are used jointly, the pair of PER and ET, which combines textual contexts and entity type information, delivers the most significant improvement.
The pair of PER and FER also performs well, which reflects the finding in \cite{peng2020learning} that context is the most important source of information.
The version with all tasks except CR sees the least drop in F1, indicating that CR's supervision signals on capturing contexts can be covered in part by PER and FER.
Last but not least, the SAIS pipeline with all four complementary tasks achieves the highest F1 score.
Similar trends are also recognized on CDR and GDA in Table~\ref{tab:bio_result}, where SAIS trained with both CR and ET (besides RE) scores higher than its single-task counterpart.

\begin{table}[t!]
\setlength\tabcolsep{2pt}
  \centering
\resizebox{\columnwidth}{!}{
  \begin{tabular}{lcc}
    \toprule
        \textbf{Model} & \textbf{CDR} & \textbf{GDA} \\
    \midrule
    \midrule
        % BRAN\text{BRAN} \cite{verga2018simultaneously} & 62.1 & - \\
        % CNN\text{CNN} \cite{nguyen2018convolutional} & 62.3 & - \\
        % $\text{EoG}$ \cite{christopoulou2019connecting} & 63.6 & 81.5 \\
        $\text{LSR}$ \cite{nan2020reasoning} & 64.8 & 82.2 \\
        $\text{SciBERT}$ \cite{beltagy2019scibert} & 65.1 & 82.5 \\
        $\text{DHG}$ \cite{zhang2020document} & 65.9 & 83.1 \\
        % $\text{GLRE}$ \cite{wang2020global} & 68.5 & - \\
        $\text{SSAN-SciBERT}$ \cite{xu2021entity} & 68.7 & 83.7 \\
        $\text{ATLOP-SciBERT}$ \cite{zhou2021document} & 69.4 & 83.9 \\
        $\text{SIRE-BioBERT}$ \cite{zeng2021sire} & 70.8 & 84.7 \\
        $\text{DocuNet-SciBERT}$ \cite{zhang2021document} & 76.3 & 85.3 \\
    \midrule
        $\text{SAIS}^\text{O}_\text{RE+CR+ET}\text{-SciBERT}$ (Ours) & \textbf{79.0} $\pm$ \textbf{0.8} & \textbf{87.1} $\pm$ \textbf{0.3} \\
    \midrule
        $\text{SAIS}^\text{O}_\text{RE+ET}\text{-SciBERT}$ & $75.9 \pm 0.9$ & $86.1 \pm 0.5$ \\
        $\text{SAIS}^\text{O}_\text{RE+CR}\text{-SciBERT}$ & $74.5 \pm 0.4$ & $85.4 \pm 0.2$ \\
        $\text{SAIS}^\text{O}_\text{RE}\text{-SciBERT}$ & $72.8 \pm 0.6$ & $ 84.5 \pm 0.3$ \\
    \bottomrule
  \end{tabular}
}
  \caption{RE F1 results ($\%$) on the CDR and GDA test sets.
  The baseline scores are from the corresponding papers.
  $\text{SAIS}_\text{RE+CR+ET}^\text{O}$ scores the highest on both datasets.
%   Baselines delivering less competitive results are omitted here.
  Full details in Appendix~\ref{ap:c}.}
  \label{tab:bio_result}
\end{table}

Moreover, as compared to the original pipeline $\text{SAIS}_\text{All}^\text{O}$, pseudo document-based data augmentation $\text{SAIS}_\text{All}^\text{D}$ acts as a hard filter by directly removing predicted non-evidence sentences, while attention mask-based data augmentation $\text{SAIS}_\text{All}^\text{M}$ distills the context more softly.
Therefore, we observe in Table~\ref{tab:ablation_ensemble} that $\text{SAIS}_\text{All}^\text{D}$ earns a higher precision, whereas $\text{SAIS}_\text{All}^\text{M}$ attains a higher recall.
By ensembling $\text{SAIS}_\text{All}^\text{O}$, $\text{SAIS}_\text{All}^\text{D}$, and $\text{SAIS}_\text{All}^\text{M}$, we improve the RE F1 score by $0.57\%$ absolutely on the DocRED develop set.

\begin{table}[t!]
  \centering
\resizebox{\columnwidth}{!}{
  \begin{tabular}{ccccc|c} 
    \toprule
        \textbf{CR} & \textbf{ET} & \textbf{PER} & \textbf{FER} & \textbf{RE} & \textbf{F1}\\
    \midrule
    \midrule
              &        &        &        & \cmark & $61.18 \pm 0.09$ \\
    \midrule
        \cmark &        &        &        & \cmark & $61.41 \pm 0.11$ \\
              & \cmark &        &        & \cmark & $61.52 \pm 0.10$ \\
              &        & \cmark &        & \cmark & $61.68 \pm 0.04$ \\
              &        &        & \cmark & \cmark & $61.44 \pm 0.07$ \\
  \midrule
        \cmark & \cmark &        &        & \cmark & $61.65 \pm 0.12$ \\
        \cmark &        & \cmark &        & \cmark & $61.79 \pm 0.08$ \\
        \cmark &        &        & \cmark & \cmark & $61.64 \pm 0.10$ \\
              & \cmark & \cmark &        & \cmark & $61.88 \pm 0.05$ \\
              & \cmark &        & \cmark & \cmark & $61.81 \pm 0.04$ \\
              &        & \cmark & \cmark & \cmark & $61.85 \pm 0.10$ \\
    \midrule
              & \cmark & \cmark & \cmark & \cmark & $62.13 \pm 0.04$ \\
        \cmark &        & \cmark & \cmark & \cmark & $62.06 \pm 0.09$ \\
        \cmark & \cmark &        & \cmark & \cmark & $61.91 \pm 0.06$ \\
        \cmark & \cmark & \cmark &        & \cmark & $61.98 \pm 0.05$ \\
    \midrule
        \cmark & \cmark & \cmark & \cmark & \cmark & $62.39 \pm 0.08$ \\
    \bottomrule
  \end{tabular}
}
  \caption{Ablation study ($\%$) using $\text{SAIS}^\text{O}\text{-BERT}_\text{base}$ to assess the effectiveness of the four complementary tasks (i.e., CR, ET, PER, and FER) for RE based on the DocRED develop set.}
  \label{tab:ablation_task}
\end{table}

% \begin{table}[t!]
%   \setlength\tabcolsep{4pt}
%   \centering
% \resizebox{\columnwidth}{!}{
%   \begin{tabular}{ccc|ccc} 
%     \toprule
%         $\textbf{SAIS}_\textbf{All}^\textbf{O} &  & \textbf{SAIS}_\textbf{All}^\textbf{D} &  & \textbf{SAIS}_\textbf{All}^\textbf{M}$ & \textbf{Precision} & \textbf{Recall} & \textbf{F1} \\
%     \midrule
%     \midrule
%         \cmark &        &        & 66.58 & 58.70 & 62.39 \\
%               & \cmark &        & 73.21 & 45.59 & 56.19 \\
%               &        & \cmark & 53.14 & 67.49 & 59.46 \\
%     % \midrule
%         % \cmark & \cmark &        & 71.14 & 54.35 & 61.62 \\
%         % \cmark &        & \cmark & 61.61 & 62.90 & 62.25 \\
%     % \midrule
%         \cmark & \cmark & \cmark & 67.76 & 58.79 & 62.96 \\
%     \bottomrule
%   \end{tabular}
% }
%   \caption{Ablation study (\%\%) using \text{BERT}_\text{base}\text{BERT}_\text{base} to assess the effectiveness of data augmentation (i.e., original ($\text{SAIS}_\text{All}^\text{O}), pseudo document-based (), pseudo document-based (\text{SAIS}_\text{All}^\text{D}), and attention mask-based (), and attention mask-based (\text{SAIS}_\text{All}^\text{M}$)) for RE based on the DocRED develop set. Full results in Appendix~?????????\ref{ap:c}.}
%   \label{tab:ablation_ensemble}
% \end{table}

\begin{table}[t!]
  \setlength\tabcolsep{4pt}
  \centering
\resizebox{\columnwidth}{!}{
  \begin{tabular}{ccc|ccc} 
    \toprule
        $\textbf{SAIS}_\textbf{All}^\textbf{O}$ & $\textbf{SAIS}_\textbf{All}^\textbf{D}$ & $\textbf{SAIS}_\textbf{All}^\textbf{M}$ & \textbf{Precision} & \textbf{Recall} & \textbf{F1} \\
    \midrule
    \midrule
        \cmark &        &        & 66.58 & 58.70 & 62.39 \\
               & \cmark &        & 73.21 & 45.59 & 56.19 \\
               &        & \cmark & 53.14 & 67.49 & 59.46 \\
    \midrule
        \cmark & \cmark &        & 71.14 & 54.35 & 61.62 \\
        \cmark &        & \cmark & 61.61 & 62.90 & 62.25 \\
    \midrule
        \cmark & \cmark & \cmark & 67.76 & 58.79 & 62.96 \\
    \bottomrule
  \end{tabular}
}
  \caption{Ablation study ($\%$) using $\text{BERT}_\text{base}$ to assess the effectiveness of data augmentation (i.e., original ($\text{SAIS}_\text{All}^\text{O}$), pseudo document-based ($\text{SAIS}_\text{All}^\text{D}$), and attention mask-based ($\text{SAIS}_\text{All}^\text{M}$)) for RE based on the DocRED develop set.}
  \label{tab:ablation_ensemble}
\end{table}

\begin{figure*}[t!]
    \centering
    \includegraphics[width=\textwidth]{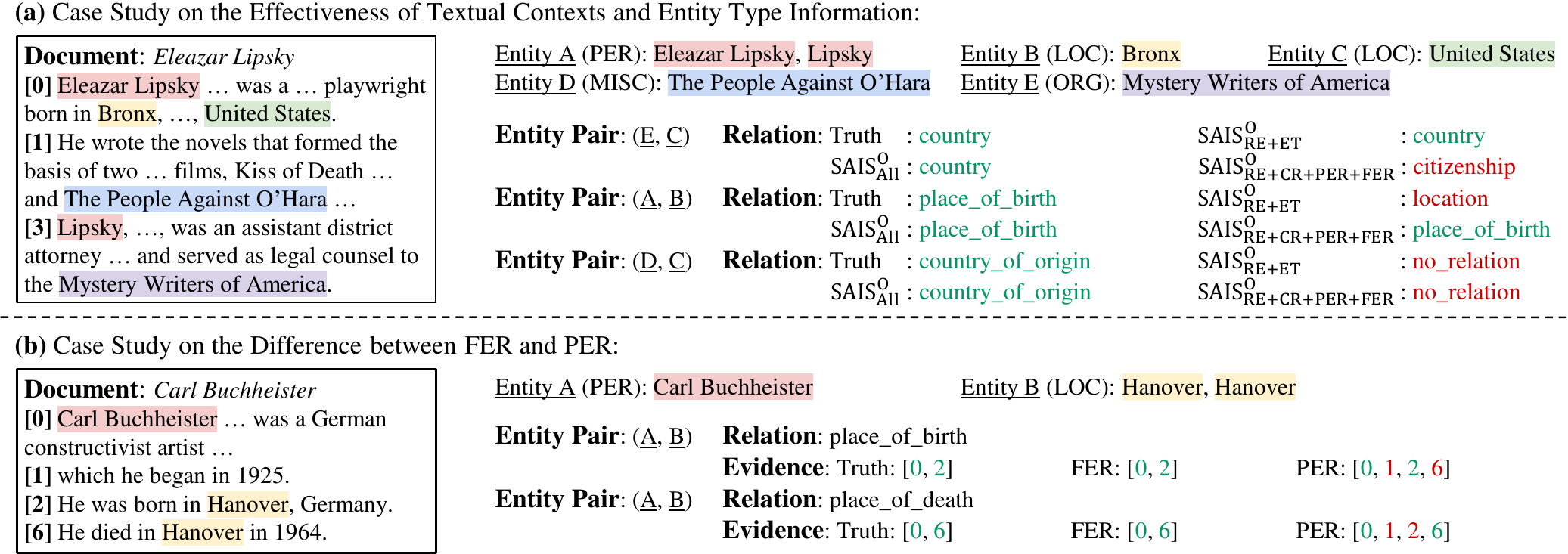}
    \caption{
    \textbf{(a)} 
    Case study on the effectiveness of textual contexts and entity type information based on models' extracted relations from the DocRED develop set.
    By capturing contexts across sentences and regularizing them with entity type information, $\text{SAIS}_\text{All}^\text{O}$ extracts relations of better quality.
    \textbf{(b)} 
    Case study on the difference between FER and PER based on retrieved evidence from the DocRED develop set. 
    FER considers evidence unique to each relation for better interpretability.
    Irrelevant sentences are omitted here.
    } 
    \label{fig:3}
\end{figure*}

\subsection{Qualitative Analysis} \label{sec:5.4}

To obtain a more insightful understanding of how textual contexts and entity type information help with RE, we present a case study in Figure~\ref{fig:3} (a).
Here, $\text{SAIS}_\text{RE+ET}^\text{O}$ is trained with the task (i.e., ET) related to entity type information while $\text{SAIS}_\text{RE+CR+PER+FER}^\text{O}$ is trained with the tasks (i.e., CR, PER, and FER) related to textual contexts.
Compared to $\text{SAIS}_\text{All}^\text{O}$, which is trained with all four complementary tasks, they both exhibit drawbacks qualitatively.
In particular, $\text{SAIS}_\text{RE+ET}^\text{O}$ can easily infer the relation ``\textit{country}" between Entities E and C based on their respective types ``\textit{ORG}" and ``\textit{LOC}", whereas $\text{SAIS}_\text{RE+CR+PER+FER}^\text{O}$ may misinterpret Entity E as of type ``\textit{PER}" and infer the relation ``\textit{citizenship}" wrongly.
On the other hand, $\text{SAIS}_\text{RE+CR+PER+FER}^\text{O}$ can directly predict the relation ``\textit{place\_of\_birth}" between Entities A and B by pattern matching, while overemphasizing the type ``\textit{LOC}" of Entity B may cause $\text{SAIS}_\text{RE+ET}^\text{O}$ to deliver the wrong relation prediction ``\textit{location}".
Last but not least, $\text{SAIS}_\text{All}^\text{O}$ effectively models contexts spanning multiple sentences and regularizes them with entity type information.
As a result, it is the only SAIS variant that correctly predicts the relation ``\textit{country\_of\_origin}" between Entities D and C.

Furthermore, to examine why SAIS (which uses FER for retrieving evidence) outperforms Eider \cite{xie2021eider} (which uses PER) significantly on ER in Table~\ref{tab:docred_result}, we compare the performance of FER and PER based on a case study in Figure~\ref{fig:3} (b).
More specifically, PER identifies the same set of evidence for both relations between Entities A and B, among which Sentence 2 describes ``\textit{place\_of\_birth}" while Sentence 6 discusses ``\textit{place\_of\_death}".
In contrast, FER considers an evidence set unique to each relation and outputs more interpretable results.
\section{Conclusion} \label{sec:6}

In this paper, we propose to explicitly teach the model to capture the major information sources of RE---textual contexts and entity types by \underline{S}upervising and \underline{A}ugmenting \underline{I}ntermediate \underline{S}teps (SAIS).
Based on a broad spectrum of carefully designed tasks, SAIS extracts relations of enhanced quality due to more effective supervision and retrieves more accurate evidence for improved interpretability.
SAIS further boosts the performance with evidence-based data augmentation and ensemble inference while preserving the computational cost by assessing model uncertainty.
Experiments on three benchmarks demonstrate the state-of-the-art performance of SAIS on both RE and ER.

If given a plain document, we shall utilize existing tools (e.g., spaCy) to get noisy annotations and apply our method afterward. 
It is also interesting to investigate how other tasks (e.g., named entity recognition) could be incorporated into the multi-task learning pipeline of our SAIS method.
We plan to explore these extensions in future works.

% \clearpage
\balance
\bibliographystyle{acl_natbib}
\bibliography{main}

\clearpage
\appendix
\onecolumn
\section{Multi-Task Learning Pipeline by Supervising Intermediate Steps ($\text{SAIS}_\text{All}^\text{O}$)} \label{ap:a}

\begin{figure*}[h!]
    \centering
    \includegraphics[width=\textwidth]{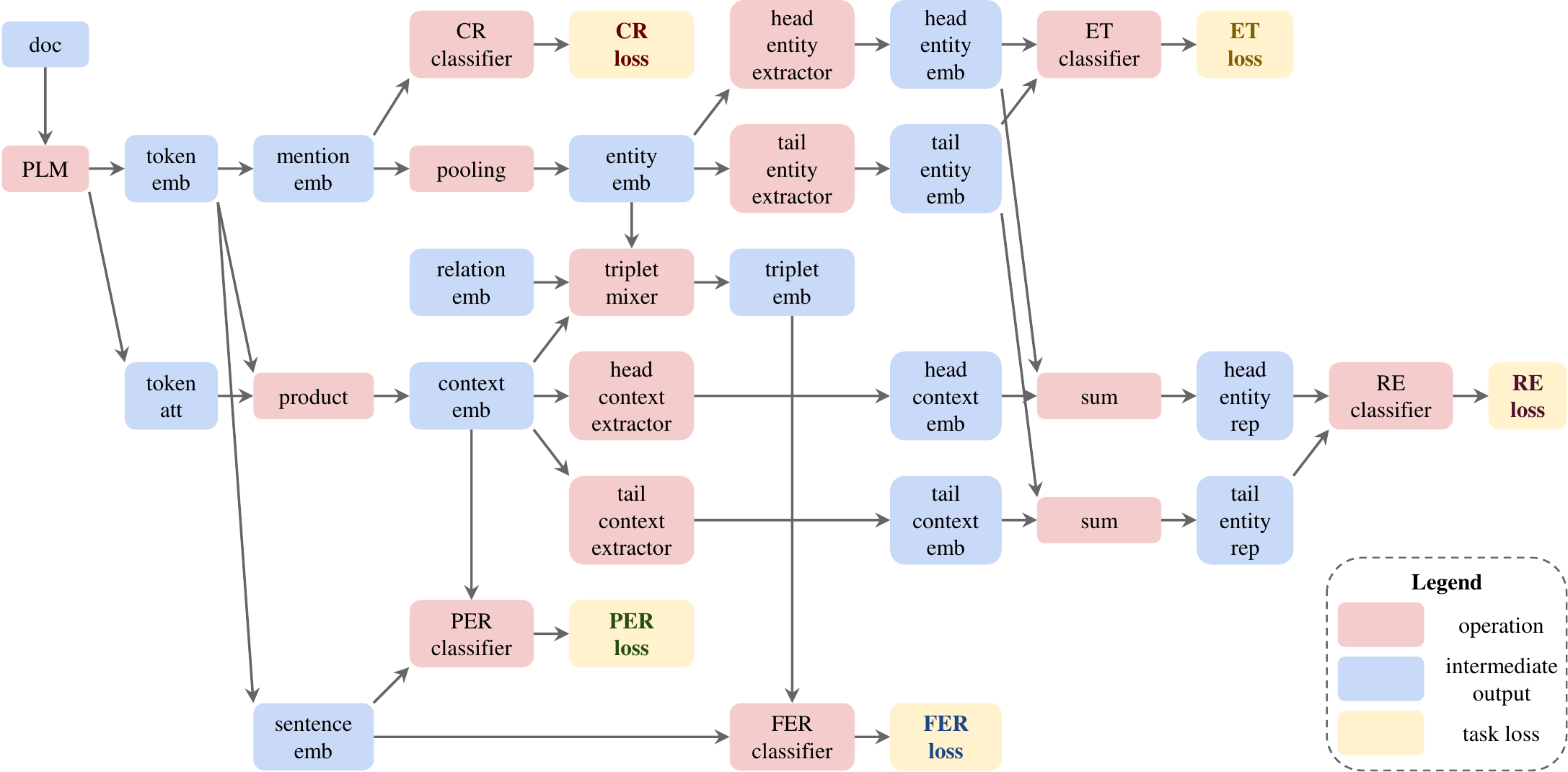}
    \caption{The overall multi-task learning pipeline of the proposed SAIS method ($\text{SAIS}_\text{All}^\text{O}$).
    By explicitly supervising the model's outputs in the intermediate steps via carefully designed tasks, we improve the RE performance.}
    \label{fig:4}
\end{figure*}

To explicitly teach the model to capture relevant contexts and entity type information for RE, we design four tasks to supervise the model's outputs in the corresponding intermediate steps.
We illustrate the overall multi-task pipeline $\text{SAIS}_\text{All}^\text{O}$ in Figure~\ref{fig:4}.

\newpage
\section{Ensemble Inference Algorithm with Evidence-based Data Augmentation ($\text{SAIS}_\text{All}^\text{B}$)} \label{ap:b}

After training the multi-task pipeline $\text{SAIS}_\text{All}^\text{O}$ proposed in Section~\ref{sec:3}, we further boost the model performance by evidence-based data augmentation and ensemble inference as discussed in Section~\ref{sec:4}. The detailed steps are explained in Algorithm~\ref{al:inference} below.

\begin{figure}[h!]
\begin{algorithm2e}[H]
\SetAlgoLined
\KwInput{trained pipeline $\text{SAIS}_\text{All}^\text{O}$ from Section~\ref{sec:3}, FER threshold $\alpha^\text{FER}$, develop set $\mathcal{D}_\text{dev}$, test set $\mathcal{D}_\text{test}$}
  \For{$\mathcal{D} \in \{\mathcal{D}_\text{dev}, \mathcal{D}_\text{test} \}$}
  {
  \begin{description}[nolistsep]
    \item [Original RE Prediction with $\text{SAIS}_\text{All}^\text{O}$ \textnormal{(Section~\ref{sec:3.6})}:] \hfill \\
        For $(e_h, e_t, r) \in \mathcal{D}$,  get $\mathbb{L}^\text{O}_{h,t,r}$ from $\text{SAIS}_\text{All}^\text{O}$. \\
    \item [Identify the Uncertain Set $\boldsymbol{\mathcal{U}}$ \textnormal{(Section~\ref{sec:4.1})}:] \hfill \\
         If $\mathcal{D}$ is $\mathcal{D}_\text{dev}$, calculate $\theta\%$ by minimizing $(\text{risk}^2 + \text{rejection rate}^2)$. \\
         $\mathcal{U}$ contains triplets with the lowest $\theta\%$ absolute confidence $|\mathbb{L}^\text{O}_{h,t,r}|$. \\
    \item [Predict Evidence Probability for $\boldsymbol{(e_h, e_t, r) \in \mathcal{U}}$ with $\text{SAIS}_\text{All}^\text{O}$ \textnormal{(Section~\ref{sec:3.5})}:] \hfill \\
         For $(e_h, e_t, r) \in \mathcal{U}$ and $s \in \mathcal{S}_d$ in the corresponding document $d$, get $\mathbb{P}_{h,t,r,s}^\text{FER}$ from $\text{SAIS}_\text{All}^\text{O}$.\\
    \item [Pseudo Document-based Data Augmentation $\text{SAIS}_\text{All}^\text{D}$ \textnormal{(Section~\ref{sec:4.2})}:] \hfill \\
         For $(e_h, e_t, r) \in \mathcal{U}$, get $\mathbb{L}^\text{D}_{h,t,r}$ by feeding the corresponding pseudo document into $\text{SAIS}_\text{All}^\text{O}$.\\
    \item [Attention Mask-based Data Augmentation $\text{SAIS}_\text{All}^\text{M}$ \textnormal{(Section~\ref{sec:4.2})}:] \hfill \\
         For $(e_h, e_t, r) \in \mathcal{U}$, get $\mathbb{L}^\text{M}_{h,t,r}$ by applying the corresponding attention mask to $\text{SAIS}_\text{All}^\text{O}$.\\
    \item [Ensemble Inference $\text{SAIS}_\text{All}^\text{B}$ \textnormal{(Section~\ref{sec:4.2})}:] \hfill \\
         If $\mathcal{D}$ is $\mathcal{D}_\text{dev}$, train $\tau_r$ for $r \in \mathcal{R}$ based on $\mathbb{L}^\text{O}_{h,t,r}$, $\mathbb{L}^\text{D}_{h,t,r}$, and $\mathbb{L}^\text{M}_{h,t,r}$ for $(e_h,e_t,r) \in \mathcal{U}$.\\
         For $(e_h, e_t, r) \in \mathcal{U}$,  get $\mathbb{L}^\text{B}_{h,t,r} = \mathbb{L}^\text{O}_{h,t,r} + \mathbb{L}^\text{D}_{h,t,r} + \mathbb{L}^\text{M}_{h,t,r} - \tau_r$.\\
    \item [Ultimate RE Prediction with $\text{SAIS}_\text{All}^\text{B}$ and $\text{SAIS}_\text{All}^\text{O}$ \textnormal{(Section~\ref{sec:4.2} and~\ref{sec:3.6})}:] \hfill \\
         For $(e_h, e_t, r) \in \mathcal{U}$, extract relation $r$ for entity pair $(e_h, e_t)$ if $\mathbb{L}^\text{B}_{h,t,r} > 0$.\\
         For $(e_h, e_t, r) \notin \mathcal{U}$, extract relation $r$ for entity pair $(e_h, e_t)$ if $\mathbb{L}^\text{O}_{h,t,r} > 0$.\\
    \item [Ultimate ER Prediction with $\text{SAIS}_\text{All}^\text{O}$ \textnormal{(Section~\ref{sec:3.5})}:] \hfill \\
         For predicted $(e_h, e_t, r)$, retrieve $s \in \mathcal{S}_d$ in the corresponding document $d$ if $\mathbb{P}_{h,t,r,s}^\text{FER} > \alpha^\text{FER}$.
  \end{description}
  }
  \KwOutput{sets of predicted triplet $(e_h, e_t, r)$ and corresponding evidence $\mathcal{V}_{h,t,r}$ for $\mathcal{D}_\text{dev}$ and $\mathcal{D}_\text{test}$}
\caption{Evidence-based Data Augmentation and Ensemble Inference ($\text{SAIS}_\text{All}^\text{B}$)}
\label{al:inference}
\end{algorithm2e}
\end{figure}
\newpage
\section{Experiment Details} \label{ap:c}

We compare the proposed SAIS method against existing baselines based on three benchmarks: CDR \cite{li2016biocreative} and GDA \cite{wu2019renet} in Table~\ref{tab:bio_result_full}, and DocRED \cite{yao2019docred} in Table~\ref{tab:docred_result_full}.
% Using the develop set of DocRED, we also evaluate the effectiveness of the four complementary tasks in~\ref{tab:ablation_task_full} and the two evidence-based data augmentation methods in~\ref{tab:ablation_ensemble_full}. 
The details are explained in Section~\ref{sec:5}.

In particular, DocRED uses the MIT License, CDR is freely available for the research community, and GDA uses the GNU Affero General Public License. 
DocRED is constructed from Wikipedia and Wikidata and, therefore, contains information that names people. 
However, since our research focuses on identifying relations among real-world entities (including public figures) based on a given document, it is impossible to fully anonymize the dataset.
We ensure that we only use publicly available information in our experiments. 
Our use of these datasets is consistent with their intended use. 
Although our method achieves state-of-the-art performance for RE and ER, using the predicted relations and evidence directly for downstream tasks without manual validation may increase the risk of errors carried forward due to the incorrect predictions.
The experiments in this paper focus on English documents from biomedical and general domains, but our proposed framework can be easily extended to documents of other languages.

\vspace{5mm}
\begin{table*}[h!]
  \centering
  \setlength\tabcolsep{3.8pt}
% \resizebox{0.575\textwidth}{!}{
  \begin{tabular}{lcc}
    \toprule
        \textbf{Model} & \textbf{CDR} & \textbf{GDA} \\
    \midrule
    \midrule
        $\text{BRAN}$ \cite{verga2018simultaneously} & 62.1 & - \\
        $\text{CNN}$ \cite{nguyen2018convolutional} & 62.3 & - \\
        $\text{EoG}$ \cite{christopoulou2019connecting} & 63.6 & 81.5 \\
        $\text{LSR}$ \cite{nan2020reasoning} & 64.8 & 82.2 \\
        $\text{SciBERT}$ \cite{beltagy2019scibert} & 65.1 & 82.5 \\
        $\text{DHG}$ \cite{zhang2020document} & 65.9 & 83.1 \\
        $\text{GLRE}$ \cite{wang2020global} & 68.5 & - \\
        $\text{SSAN-SciBERT}$ \cite{xu2021entity} & 68.7 & 83.7 \\
        $\text{ATLOP-SciBERT}$ \cite{zhou2021document} & 69.4 & 83.9 \\
        $\text{SIRE-BioBERT}$ \cite{zeng2021sire} & 70.8 & 84.7 \\
        $\text{DocuNet-SciBERT}$ \cite{zhang2021document} & 76.3 & 85.3 \\
    \midrule
        $\text{SAIS}^\text{O}_\text{RE+CR+ET}\text{-SciBERT}$ (Ours) & \textbf{79.0} $\pm$ \textbf{0.8} & \textbf{87.1} $\pm$ \textbf{0.3} \\
    \midrule
        $\text{SAIS}^\text{O}_\text{RE+ET}\text{-SciBERT}$ & $75.9 \pm 0.9$ & $86.1 \pm 0.5$ \\
        $\text{SAIS}^\text{O}_\text{RE+CR}\text{-SciBERT}$ & $74.5 \pm 0.4$ & $85.4 \pm 0.2$ \\
        $\text{SAIS}^\text{O}_\text{RE}\text{-SciBERT}$ & $72.8 \pm 0.6$ & $84.5 \pm 0.3$ \\
    \bottomrule
  \end{tabular}
% }
  \caption{RE F1 results ($\%$) on the CDR and GDA test sets.
  We report the baseline performances from the corresponding papers.
  $\text{SAIS}_\text{RE+CR+ET}^\text{O}$ using three training tasks (i.e., RE, CR, and ET) scores the highest on both datasets and better than its variants with fewer training tasks.}
  \label{tab:bio_result_full}
\end{table*}

\newpage

\begin{table*}[t!]
  \centering
  \setlength\tabcolsep{3.8pt}
\resizebox{\textwidth}{!}{
  \begin{tabular}{lcccccc} 
    \toprule
         & \multicolumn{3}{c}{\textbf{DocRED Dev}} & \multicolumn{3}{c}{\textbf{DocRED Test}} \\
    \cmidrule(lr){2-4} \cmidrule(lr){5-7}
        \textbf{Model} & \multicolumn{2}{c}{Relation} & Evidence & \multicolumn{2}{c}{Relation} & Evidence \\
    \cmidrule(lr){2-3} \cmidrule(lr){4-4} \cmidrule(lr){5-6} \cmidrule(lr){7-7}
        & Ign F1 & F1 & F1 & Ign F1 & F1 & F1 \\
    \midrule
    \midrule
        $\text{CNN}$ \cite{yao2019docred} & 41.58 & 43.45 & - & 40.33 & 42.26 & - \\
        $\text{GAT}$ \cite{velivckovic2017graph} & 45.17 & 51.44 & - & 47.36 & 49.51 & - \\
        $\text{BiLSTM}$ \cite{yao2019docred} & 48.87 & 50.94 & 44.07 & 48.78 & 51.06 & 43.83 \\
        $\text{GCNN}$ \cite{sahu2019inter} & 46.22 & 51.52 & - & 49.59 & 51.62 & - \\
        $\text{EoG}$ \cite{christopoulou2019connecting} & 45.94 & 52.15 & - & 49.48 & 51.82 & - \\
        $\text{AGGCN}$ \cite{guo2019attention} & 46.29 & 52.47 & - & 48.89 & 51.45 & - \\
    \midrule
    \midrule
        $\text{GEDA-BERT}_\text{base}$ \cite{li2020graph} & 54.52 & 56.16 & - & 53.71 & 55.74 & - \\
        $\text{GLRE-BERT}_\text{base}$ \cite{wang2020global} & - & - & - & 55.40 & 57.40 & - \\
        $\text{LSR-BERT}_\text{base}$ \cite{nan2020reasoning} & 52.43 & 59.00 & - & 56.97 & 59.05 & - \\
        $\text{HeterGSAN-BERT}_\text{base}$ \cite{xu2021document} & 58.13 & 60.18 & - & 57.12 & 59.45 & - \\
        $\text{GAIN-BERT}_\text{base}$ \cite{zeng2020double} & 59.14 & 61.22 & - & 59.00 & 61.24 & - \\
        $\text{DRN-BERT}_\text{base}$ \cite{xu2021discriminative} & 59.33 & 61.39 & - & 59.15 & 61.37 & - \\
        $\text{SIRE-BERT}_\text{base}$ \cite{zeng2021sire} & 59.82 & 61.60 & - & 60.18 & 62.05 & - \\ 
    \midrule
        $\text{BERT}_\text{base}$ \cite{wang2019fine} & - & 54.16 & - & - & 53.20 & - \\
        $\text{BERT-TS}_\text{base}$ \cite{wang2019fine} & - & 54.42 & - & - & 53.92 & - \\
        $\text{HIN-BERT}_\text{base}$ \cite{tang2020hin} & 54.29 & 56.31 & - & 53.70 & 55.60 & - \\
        $\text{CorefBERT}_\text{base}$ \cite{ye2020coreferential} & 55.32 & 57.51 & - & 54.54 & 56.96 & - \\
        $\text{E2GRE-BERT}_\text{base}$ \cite{huang2020entity} & 55.22 & 58.72 & 47.14 & - & - & - \\
        $\text{SSAN-BERT}_\text{base}$ \cite{xu2021entity} & 57.03 & 59.19 & - & 56.06 & 58.41 & - \\
        $\text{ATLOP-BERT}_\text{base}$ \cite{zhou2021document} & 59.22 & 61.09 & - & 59.31 & 61.30 & - \\
        $\text{DocuNet-BERT}_\text{base}$ \cite{zhang2021document} & 59.86 & 61.83 & - & 59.93 & 61.86 & - \\
        $\text{Eider-BERT}_\text{base}$ \cite{xie2021eider} & \textbf{60.51} & 62.48 & 50.71 & 60.42 & 62.47 & 51.27 \\
    \midrule
        $\text{SAIS}^\text{B}_\text{All}\text{-BERT}_\text{base}$ (Ours) & 59.98 $\pm$ 0.13 & \textbf{62.96} $\pm$ \textbf{0.11} & \textbf{53.70} $\pm$ \textbf{0.21} & \textbf{60.96} & \textbf{62.77} & \textbf{52.88} \\
    \midrule
    \midrule
        $\text{BERT}_\text{large}$ \cite{ye2020coreferential} & 56.51 & 58.70 & - & 56.01 & 58.31 & - \\
        $\text{CorefBERT}_\text{large}$ \cite{ye2020coreferential} & 56.82 & 59.01 & - & 56.40 & 58.83 & - \\
        $\text{RoBERTa}_\text{large}$ \cite{ye2020coreferential} & 57.19 & 59.40 & - & 57.74 & 60.06 & - \\
        $\text{CorefRoBERTa}_\text{large}$ \cite{ye2020coreferential} & 57.35 & 59.43 & - & 57.90 & 60.25 & - \\
        $\text{SSAN-RoBERTa}_\text{large}$ \cite{xu2021entity} & 60.25 & 62.08 & - & 59.47 & 61.42 & - \\
        $\text{E2GRE-RoBERTa}_\text{large}$ \cite{huang2020entity} & - & - & - & 60.30 & 62.50 & 50.50 \\
        $\text{ATLOP-RoBERTa}_\text{large}$ \cite{zhou2021document} & 61.32 & 63.18 & - & 61.39 & 63.40 & - \\
        $\text{DocuNet-RoBERTa}_\text{large}$ \cite{zhang2021document} & 62.23 & 64.12 & - & 62.39 & 64.55 & - \\
        $\text{Eider-RoBERTa}_\text{large}$ \cite{xie2021eider} & \textbf{62.34} & 64.27 & 52.54 & 62.85 & 64.79 & 53.01 \\
    \midrule
        $\text{SAIS}^\text{B}_\text{All}\text{-RoBERTa}_\text{large}$ (Ours) & 62.23 $\pm$ 0.15 & \textbf{65.17} $\pm$ \textbf{0.08} & \textbf{55.84} $\pm$ \textbf{0.23} & \textbf{63.44} & \textbf{65.11} & \textbf{55.67} \\
    \bottomrule
  \end{tabular}
}
  \caption{RE and ER results ($\%$) on the develop and test sets of DocRED. 
  Ign F1 refers to the F1 score excluding the relation instances mentioned in the train set.
  Baselines using $\text{BERT}_\text{base}$ are separated into the graph-based (upper) and transformer-based (lower) groups. 
  We report the test set scores from the official scoreboard and the baseline scores from the corresponding papers.
  $\text{SAIS}_\text{All}^\text{B}$ achieves state-of-the-art performance on both RE and ER.}
  \label{tab:docred_result_full}
\end{table*}

\end{document}